\documentclass[letterpaper]{article} 
\usepackage{aaai24}  
\usepackage{times}  
\usepackage{helvet}  
\usepackage{courier}  
\usepackage[hyphens]{url}  
\usepackage{graphicx} 
\usepackage{natbib}  
\usepackage{caption} 
\usepackage{algorithm}
\usepackage{algorithmic}

\usepackage{newfloat}
\usepackage{listings}
\DeclareCaptionStyle{ruled}{labelfont=normalfont,labelsep=colon,strut=off} 
\floatstyle{ruled}
\newfloat{listing}{tb}{lst}{}
\floatname{listing}{Listing}
\usepackage{soul}
\usepackage{amsmath}
\usepackage{amsfonts}
\usepackage{amsthm}
\usepackage{booktabs}
\usepackage{xspace}
\usepackage{amssymb}
\usepackage{subfigure}
\usepackage{todonotes}
\usepackage{pifont}
\usepackage{mathtools}
\usepackage{environ}
\usepackage{multirow}
\usepackage{pdfpages}

\newcommand{\fullmodel}{\textbf{M}ulti-visit health \textbf{S}tatus \textbf{I}nference model for \textbf{C}ollaborative EHR synthesis\xspace}
\newcommand{\model}{MSIC\xspace}

\DeclareMathSymbol{\shortminus}{\mathbin}{AMSa}{"39}

\title{Collaborative Synthesis of Patient Records through\\ Multi-Visit Health State Inference}
\author{
    Hongda Sun,\textsuperscript{\rm 1,}\equalcontrib \ 
    Hongzhan Lin,\textsuperscript{\rm 1,}\equalcontrib \
    Rui Yan\textsuperscript{\rm 1,}\textsuperscript{\rm 2,}\thanks{Corresponding author: Rui Yan (ruiyan@ruc.edu.cn) }
}

\affiliations{
    \textsuperscript{\rm 1} Gaoling School of Artificial Intelligence, Renmin University of China\\
    \textsuperscript{\rm 2} Engineering Research Center of Next-Generation Intelligent Search and Recommendation, Ministry of Education\\
    \{sunhongda98, linhongzhan, ruiyan\}@ruc.edu.cn
}

\usepackage{bibentry}

\begin{document}

\maketitle

\begin{abstract}
Electronic health records (EHRs) have become the foundation of machine learning applications in healthcare, while the utility of real patient records is often limited by privacy and security concerns. Synthetic EHR generation provides an additional perspective to compensate for this limitation.
Most existing methods synthesize new records based on real EHR data, without consideration of different types of events in EHR data, which cannot control the event combinations in line with medical common sense.
In this paper, we propose \textbf{\model},  a \fullmodel to address these limitations.
First, we formulate the synthetic EHR generation process as a probabilistic graphical model and tightly connect different types of events by modeling the latent health states. Then, we derive a health state inference method tailored for the multi-visit scenario to effectively utilize previous records to synthesize current and future records.
Furthermore, we propose to generate medical reports to add textual descriptions for each medical event,  providing broader applications for synthesized EHR data.
For generating different paragraphs in each visit, we incorporate a multi-generator deliberation framework to collaborate the message passing of multiple generators and employ a two-phase decoding strategy to generate high-quality reports.
Our extensive experiments on the widely used benchmarks, MIMIC-III and MIMIC-IV, demonstrate that \model advances state-of-the-art results on the quality of synthetic data while maintaining low privacy risks.
\end{abstract}

\begin{figure}
\centering
\includegraphics[width=0.95\linewidth]{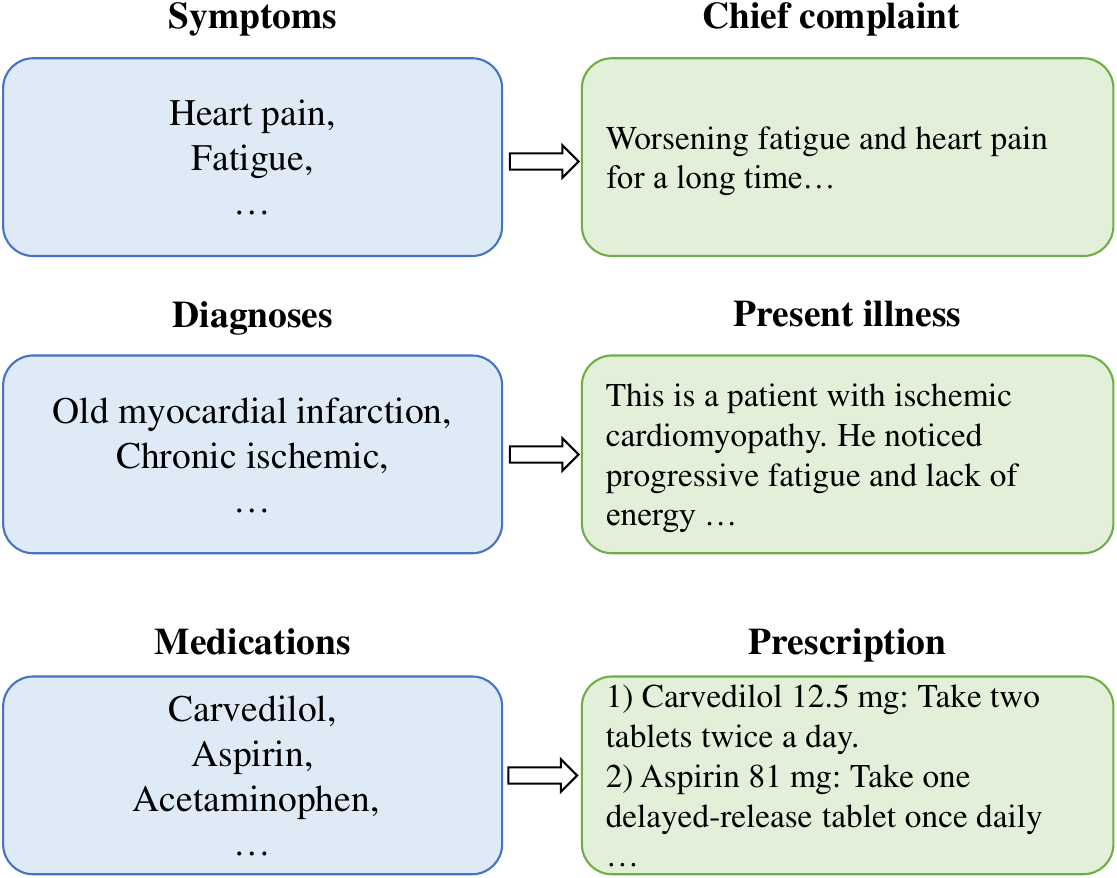}
\caption{An example of a single-visit EHR data.}
\label{fig:data}
\end{figure}

\section{Introduction}
Electronic health records (EHRs) have received extensive attention in research and applications in the AI-empowered healthcare field. Mainstream healthcare-related tasks such as medical named entity recognition~\cite{named_entity_recognition3,named_entity_recognition1,named_entity_recognition2} and drug recommendation~\cite{drug_recommendation1,drug_recommendation2,drug_recommendation3}
can be carried out based on EHR datasets. However, EHR data sharing is generally limited by privacy/policy regulations, resulting in real records being insufficient for requirements~\cite{pp1,pp2,pp3,privacy3}.

Therefore, synthetic EHR generation is a safer alternative for privacy protection~\cite{CONAN,SynTEG}.
With the promising development of deep generative models~\cite{deep_generative_model1,deep_generative_model2}, generative adversarial networks (GANs) and variational autoencoders (VAEs) are widely used for synthetic EHR generation~\cite{synthesizing_EHR_data1,synthesizing_EHR_data2,synthesizing_EHR_data4}.
These methods learn from the real EHR data to synthesize a series of new records.
However, longitudinal EHR data has a patient-visit-event hierarchical structure: (1) each patient has sequential records for multiple visits; (2) each patient visit contains various types of medical events including symptoms, diagnoses, and medications.
Previous EHR synthesis methods usually fail to model the logical associations between different types of events. Instead, they generate symptoms, diagnoses, and medications symmetrically, which cannot ensure that their combination conforms to medical common sense, affecting the quality and reliability of synthesized data.
For multi-visit synthetic scenarios, how to use previous records to accurately generate current events is also a key challenge for this task.
Furthermore, original EHR synthesis methods only focus on generating discrete labels for each medical event, and this limited medical information directly affects their application scope.

To this end, we propose a \fullmodel (\model) to address these limitations.
First, we formulate the synthetic EHR generation process as a probabilistic graphical model to capture the correlations between different types of medical events.
To better model the generation process, we define a latent health state variable for each patient as a key connection with different events.
Specifically, we treat symptoms and diagnoses as the patient's inherent factors, which determine the distribution of the patient's health state. 
Then, by inferring the health state, we can obtain appropriate medications for the patient, and generate symptoms and diagnoses in the next visit.
Second, we generalize the graphical model to the multi-visit scenario by linking the connections between events with the same type.
Correspondingly, we link the health states for each patient to make it continuously evolve across multiple visits.
Thus, we can generate the current synthetic record based on the patient's previous records and enable sequential synthesis.

Traditional EHR synthesis generates limited medical event labels, restricting task coverage.
We then propose to generate medical reports to expand the application scenarios to serve the tasks that labels cannot support, like symptom extraction~\cite{sympgraph}, diagnosis prediction~\cite{diagnosis}, and clinical outcome prediction~\cite{naik2022literature}.
An example of a single-visit EHR data is shown in Figure~\ref{fig:data}, with three types of medical events on the left and the corresponding medical report paragraphs detailing the events on the right.
The patient's health state acts as also a conditional signal to control the report generation process.
We develop a new multi-generator deliberation framework with two-phase decoding to collaborate the paragraph generation among multiple generators.
In the draft phase, we sequentially draft three paragraphs through forward message passing.
In the polish phase, we use the paragraphs drafted by each generator as deliberation feedback to refine the other two paragraphs.

We conduct extensive experiments on two benchmark datasets, MIMIC-III and MIMIC-IV, to evaluate the quality and privacy of synthetic EHR data. Experimental results demonstrate that our \model outperforms state-of-the-art baselines on both medical event synthesis and report generation. Meanwhile, evaluation results of membership and attribute inference attacks show that \model can also maintain low privacy risks for safer synthesis.

Our technical contributions are summarized as follows:

\noindent $\bullet$ We construct a probabilistic graphical model to formulate the synthetic EHR generation process. We introduce a latent health state variable for each patient to connect the logical associations between various types of medical events.

\noindent $\bullet$ We derive a new sequential health state inference method to better leverage a patient's previous visits to synthesize current and future EHR data more coherently.

\noindent $\bullet$ We propose to generate medical reports to detail each medical event using textual descriptions, providing broader applications for synthetic data. We develop a multi-generator deliberation framework and a two-phase decoding strategy to improve the collaboration of multi-paragraph generation.

\section{Related Work}
\subsection{Synthetic EHR generation}
Deep generative models have advanced promising performance on synthetic EHR generation.
GAN-style models are also widely applied in this field. MedGAN combined an autoencoder and GAN to generate high-dimensional medical codes~\cite{medGAN}.
MedGAN was improved into two more effective variants, MedWGAN and MedBGAN, to achieve comparable performance on distribution statistics~\cite{MedWGANMedBGAN}. LongGAN combined RNN and Wasserstein GAN Gradient Penalty to generate time-series data~\cite{LongGAN}. However, GAN-style methods have several drawbacks such as training instability and mode collapse~\cite{mode1,mode2}.
EVA used a variational autoencoder modeling longitudinal patterns across visits to produce records~\cite{EVA}. 
Language models can also be used in generating synthetic EHR data. PromptEHR uses prompt learning for conditional generation of synthetic data~\cite{promptehr}.
Different from previous methods, we model the logical associations between various types of medical events with multi-visit health state inference.

\subsection{Medical language modeling and generation}
Language models have been widely used in the clinical and medical fields~\cite{biomedical}. 
BioBERT pre-trained BERT with biomedical corpora for natural language understanding tasks~\cite{biobert}. ClinicalBERT~\cite{clinicalbert} and BlueBERT~\cite{BlueBERT} are also BERT-based models with more corpora.
GENRE~\cite{GENRE} and BARTNER~\cite{BARTNER} are all auto-regressive models for named entity recognition and entity linking tasks.
BioBART is pre-trained on biomedical literature and can be used for downstream generation tasks including medical dialogue and summarization~\cite{biobart}. 
We employ BioBART as the backbone of medical report generation and further collaborate with multiple generators to enhance message passing between them, improving the quality of generated reports.

\section{Preliminary}

\subsection{Problem formulation}\label{task}
An EHR dataset containing $N$ patients can be denoted as $\mathbf{X} = \{\mathbf{x}^{(1)}, \mathbf{x}^{(2)}, \cdots, \mathbf{x}^{(N)} \}$. The $j$-th patient can be characterized by a sequence of multiple visits: $\mathbf{x}^{(j)} = [\mathbf{x}^{(j)}_1, \mathbf{x}^{(j)}_2, \cdots, \mathbf{x}^{(j)}_{T_j}]$, where $T_j$ is the total number of visits of patient $j$.
Within each visit, the record encompasses various medical event types.
Specifically, the $t$-th visit of patient $j$ can be represented as $\mathbf{x}^{(j)}_t = (\mathbf{s}^{(j)}_t, \mathbf{d}^{(j)}_t, \mathbf{m}^{(j)}_t)$, where $\mathbf{s}^{(j)}_t \in [0,1]^{|\mathcal{S}|}$, $\mathbf{d}^{(j)}_t \in [0,1]^{|\mathcal{D}|}$, and $\mathbf{m}^{(j)}_t \in [0,1]^{|\mathcal{M}|}$ are binary labels of symptoms, diagnoses and medications, respectively.
Here, $\mathcal{S}$, $\mathcal{D}$, and $\mathcal{M}$ indicates the overall symptom, diagnosis, and medication sets, and $|\cdot|$ indicates the cardinality of a given set. 

The objective of synthetic EHR generation can be divided into two primary processes.
First, we aim to synthesize new accurate records from existing partial records.
In one case, \textbf{cross-type synthesis} needs to generate other types of events based on observed event types (e.g., generate $\mathbf{m}^{(j)}_t$ given $\mathbf{s}^{(j)}_t$ and $\mathbf{d}^{(j)}_t$). 
In another case, \textbf{longitudinal synthesis} needs to generate the next visit $\mathbf{x}^{(j)}_{t+1}$ based on the existing visits $(\mathbf{x}^{(j)}_1, \cdots, \mathbf{x}^{(j)}_t)$.
Second, beyond synthesizing discrete event labels, 
we introduce a supplementary task, namely \textbf{medical report generation}, to improve the informativeness of synthetic data.
Consistent with the event labels $\mathbf{x}^{(j)}_t$, the medical report $\mathbf{y}^{(j)}_t$ consists of three paragraphs: (1) \textit{chief complaint} $\mathbf{y}^{(j)}_{s_t}$ describes the patient's self-reported symptoms, corresponding to $\mathbf{s}^{(j)}_t$; (2) \textit{present illness} $\mathbf{y}^{(j)}_{d_t}$ specifies the current detailed diagnosis of the patient, corresponding to $\mathbf{d}^{(j)}_t$; and (3) \textit{prescription} $\mathbf{y}^{(j)}_{m_t}$ details the recommended medications for the patient, corresponding to $\mathbf{m}^{(j)}_t$.

\begin{figure}
\centering
\includegraphics[width=0.95\linewidth]{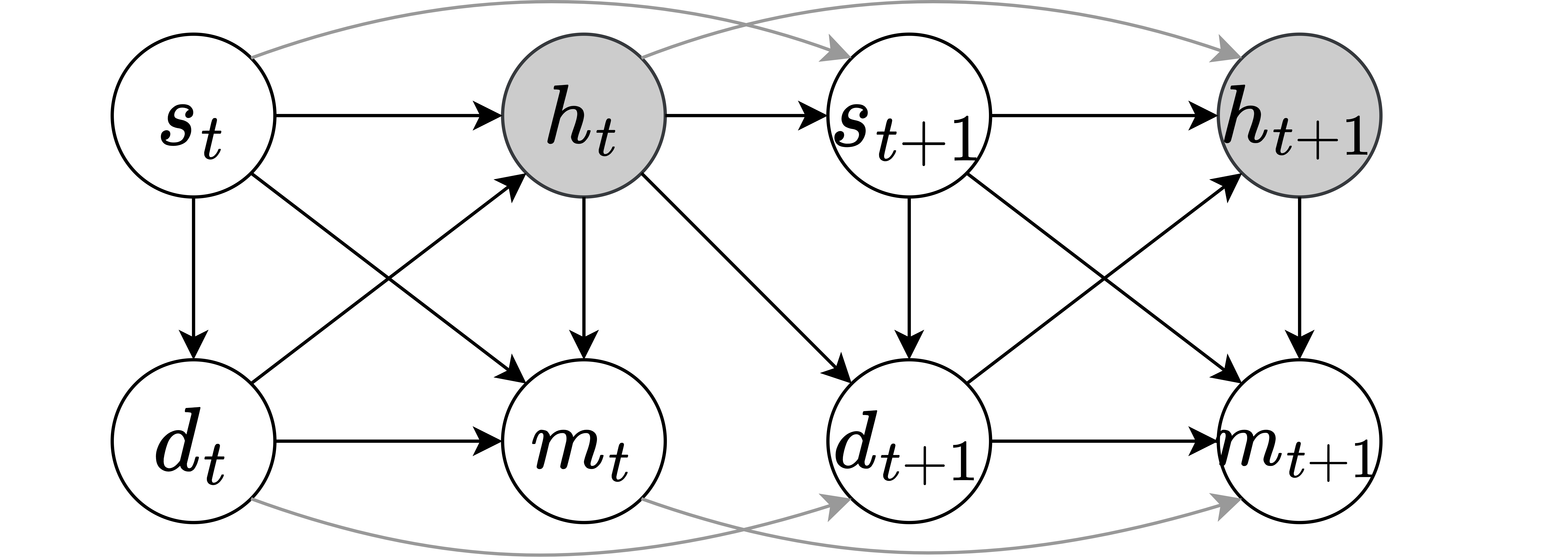}
\caption{The graphical representation of synthetic EHR generation. Shaded circles represent latent variables.}
\label{fig:g1}
\end{figure}

\subsection{Graphical model for synthetic EHR generation}
Previous synthetic EHR generation methods overlook the unique roles of different medical events and treat them as symmetrical modeling positions~\cite{EVA,medGAN}. To overcome this limitation, we consider the formation logic of real-world EHR data and design a probabilistic graphical model to capture the generative process, displayed in Figure~\ref{fig:g1}. Note that we will omit the superscript $j$ referring to a specific patient for simplification.
The process begins with establishing the diagnoses $d_t$ from observed symptoms $s_t$.
Since $s_t$ and $d_t$ are inherent factors related to the patient's health condition, we derive a health state variable $h_t$ as a crucial conditional signal in generating these events. 
The medications $m_t$ can be determined by inferring $h_t$ in conjunction with $s_t$ and $d_t$ to simulate the real-world prescription process.
Additionally, the current $h_t$ also directly influences symptoms $s_{t+1}$ and diagnoses $d_{t+1}$ in the next visit.
To enable longitudinal synthesis across visits, we expand the graphical model to accommodate multi-visit scenarios.
We model the correlations exhibited by these events by adding critical links in the graphical model.
Figure~\ref{fig:g1} illustrates the cross-type relation paths with black arrows, which are mainly centered around the health state to depict the interplay between symptoms, diagnoses, and medications.
The gray arrows represent longitudinal relation paths demonstrating the dependencies between consecutive visits, enriching the temporal dynamics modeling in synthesizing EHR data.

\section{Our Method}

\subsection{Sequential health state inference}
Following the formation logic of EHR data, we propose a new variational Bayesian generative model to formulate the generative process.
The latent health state $h_t$ is inferred and evolved over multiple visits and used for a conditional signal for both medical event synthesis and report generation.
During the $t$-th visit, the observed factors $s_t$ and $d_t$ play a crucial role in deriving $h_t$ since they directly reflect the distribution of the patient's health condition. 
Meanwhile, the patient's health state progresses sequentially across multiple visits,
necessitating consideration of the impact of the previous health state $h_{t-1}$ on the distribution of $h_t$.
Therefore, how to accurately formulate the distribution of $h_t$ is crucial for both cross-type and longitudinal EHR synthesis.
To infer the health state $h_t$, we integrate the effects of both current patient inherent factors $(s_t, d_t)$ and previous health state $h_{t-1}$, and derive its prior distribution $P(h_t | h_{t-1}, s_t, d_t)$ parameterized by a prior network $P_\theta$.
Then $h_t$ can be drawn based on the prior for generating current medications $m_t$.
To refine the inference process, we introduce the ground-truth medications $m_t$ as an additional auxiliary condition and use the posterior distribution $P(h_t | h_{t-1}, s_t, d_t, m_t)$ to estimate $h_t$. 
We approximate the posterior using an inference network $P_\phi$ due to complex expectation manipulations of the exact posterior.
Overall, we derive the inferred health state $h_t$ from the inference network $P_\phi$ in the training stage, or obtain $h_t$ from the prior network $P_\theta$ in the testing stage.
To optimize both $P_\theta$ and $P_\phi$, we derive the variational evidence lower bound (ELBO), as detailed in the following subsections.

\subsection{Synthetic EHR generation}\label{ehrgen}
Based on the inferred health state $h_t$, it serves as a conditional signal to generate synthetic EHR data. We divide synthetic EHR generation into the following two steps: (1) multi-visit synthesis for medical events labels and (2) multi-generator deliberation for medical report generation.

\begin{figure*}
    \centering
    \includegraphics[width=0.8\textwidth]{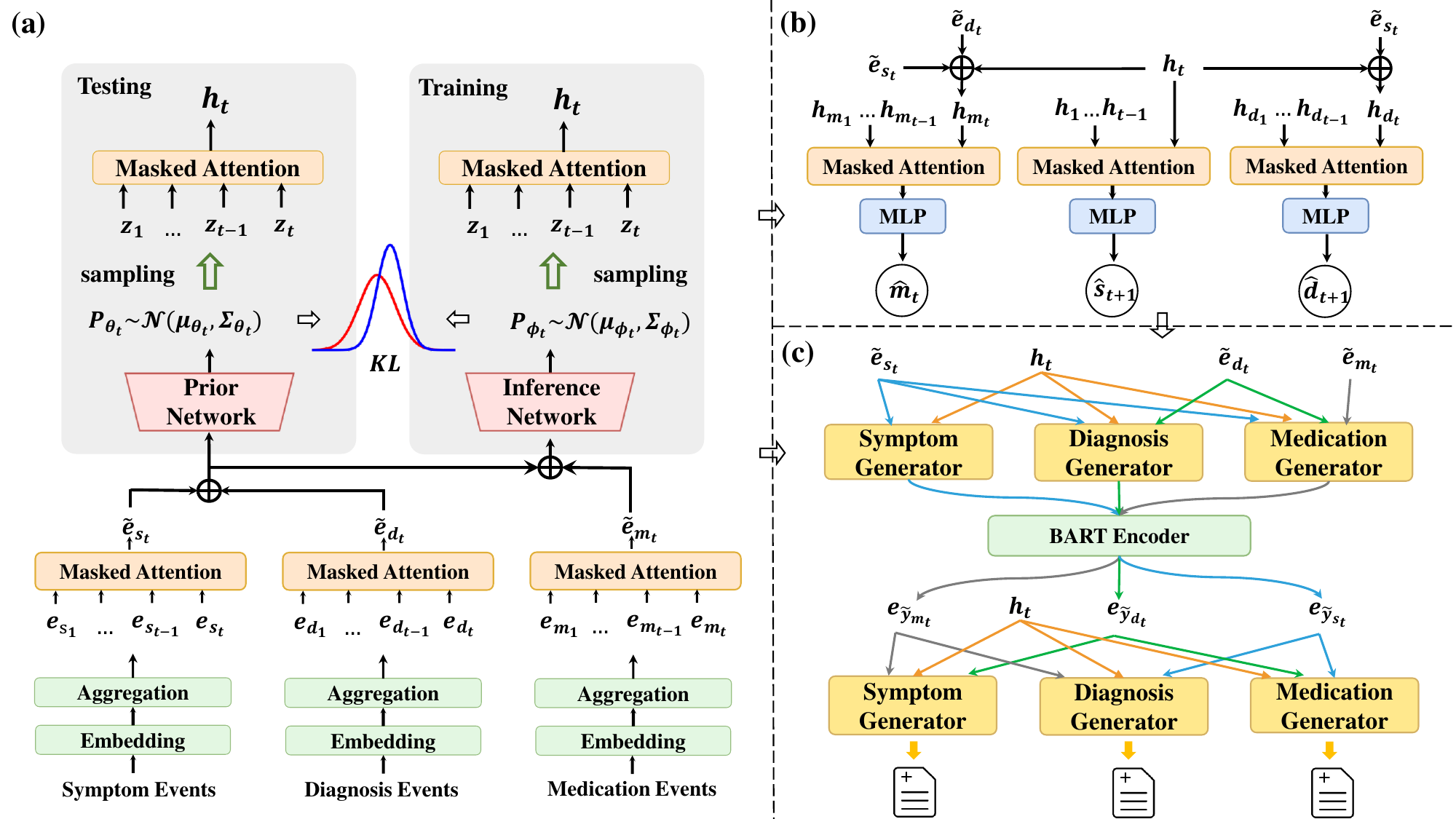}
    \caption{The model overview of \model with three modules: (a) health state inference module; (b) medical event synthesis module; (c) medical report generation module.}
    \label{fig:model}
\end{figure*}

\paragraph{Multi-visit synthesis for medical events.}
Given the previous records $\mathbf{x}_{t-1}$ and the current patient intrinsic factors $(s_t, d_t)$, we can perform cross-type synthesis for current medications by estimating $P(m_t | \mathbf{x}_{t-1}, s_t, d_t)$, 
the probability of observing each $m_t$ in $\mathbf{x}_t$.
Figure~\ref{fig:g1} indicates that we need to use the previous health state $h_{t-1}$ as a key condition of inferring the current $h_t$. Then, we can use the derived $h_{t-1}$ to generate $m_t$.
From the perspective of ELBO, we can approximate the health state distributions and derive the following probability estimation:
{\fontsize{8.5}{0}\selectfont
\begin{align}
    & \log \widehat{P}(m_t | \mathbf{x}_{t\shortminus1}, s_t, d_t)) \nonumber \\ 
 = & \ \mathbb{E}_{P_\phi(h_{t\shortminus1})}\left[\log \mathbb{E}_{P_\theta(h_t)} P(m_t | m_{t\shortminus1}, h_t, s_t, d_t)\right] \nonumber \\
 \ge & \ \mathbb{E}_{P_\phi(h_{t\shortminus1})}\left[
 \mathbb{E}_{P_\phi(h_t)}  \log g_m(m_t) -\text{KL}(P_\phi(h_t)||P_\theta(h_t))\right],\label{mt}
\end{align}
}
where $\mathbb{E}$ indicates the expectation operation, $g_m$ is the medication decoder for generating $m_t$, and KL denotes the Kullback-Leibler divergence.
$P_\phi(h_t)$ and $P_\theta(h_t)$ are the abbreviations of prior and inference networks, omitting conditionals for simplification.
The first term is to reconstruct $m_t$ from the distribution $P(m_t | m_{t-1}, h_t, s_t, d_t)$, and the second KL term is to shrink the distance between $P_\phi(h_t)$ to $P_\theta(h_t)$.
Similar to the synthesis of medications $m_t$, $h_t$ can also be used for cross-type synthesis on the symptoms $s_{t+1}$ and diagnoses $d_{t+1}$ in the next visit.
Given the current record $\mathbf{x}_t$ along with inferred $h_t$, we can derive the estimated probabilities as:
{\fontsize{8.5}{0}\selectfont
\begin{align}
 & \log \widehat{P}(s_{t+1} | \mathbf{x}_t)) \nonumber \\
 = & \ \mathbb{E}_{P_\phi(h_{t\shortminus1})}\left[\log \mathbb{E}_{P_\theta(h_t)} P(s_{t+1} | s_t, h_t)\right] \nonumber \\
 \ge & \ \mathbb{E}_{P_\phi(h_{t\shortminus1})}\left[
 \mathbb{E}_{P_\phi(h_t)}  \log g_s(s_{t+1}) -\text{KL}(P_\phi(h_t)||P_\theta(h_t))\right], 
 \label{st}
 \\
 & \log \widehat{P}(d_{t+1} | \mathbf{x}_t, s_{t+1})) \nonumber \\
 = & \ \mathbb{E}_{P_\phi(h_{t\shortminus1})}\left[\log \mathbb{E}_{P_\theta(h_t)} P(d_{t+1} | d_t, h_t, s_{t+1})\right] \nonumber \\
 \ge & \ \mathbb{E}_{P_\phi(h_{t\shortminus1})}\left[
 \mathbb{E}_{P_\phi(h_t)}  \log g_d(d_{t+1}) -\text{KL}(P_\phi(h_t)||P_\theta(h_t))\right].
 \label{dt}
\end{align}
}
where $g_s$ and $g_d$ are symptom and diagnosis decoders for generating $s_{t+1}$ and $d_{t+1}$, respectively. Therefore, we can synthesize all types of medical events given the existing partial records.
Building upon the above cross-type synthesis, we can extend our method to make longitudinal synthesis for the next visit $\mathbf{x}_{t+1}$ based on the current record $\mathbf{x}_t$. 
To achieve this, we intuitively decompose the process of $\mathbf{x}_{t+1}$ into separate sequential synthesis of $s_{t+1}$, $d_{t+1}$, and $m_{t+1}$.
Based on Equations~\eqref{mt}-\eqref{dt}
The joint probability can be decomposed and rearranged as:
{\scriptsize
\begin{align}
   & \log \widehat{P}(\mathbf{x}_{t+1} | \mathbf{x}_{t})
   \nonumber \\
 = & \log \left[\widehat{P}(s_{t+1} | \mathbf{x}_t))\cdot \widehat{P}(d_{t+1} | \mathbf{x}_t, s_{t+1}))\cdot \widehat{P}(m_{t+1} | \mathbf{x}_t, s_{t+1}, d_{t+1})) \right] \nonumber \\
 \ge & \ \mathbb{E}_{P_\phi(h_{t\shortminus1})}\left[
 \mathbb{E}_{P_\phi(h_t)}  \log [g_s(s_{t+1})g_d(d_{t+1})] \shortminus\text{KL}(P_\phi(h_t)||P_\theta(h_t))\right] \nonumber \\
 + & \ \mathbb{E}_{P_\phi(h_t)}\left[
 \mathbb{E}_{P_\phi(h_{t+1})}  \log g_m(m_{t+1}) \shortminus\text{KL}(P_\phi(h_{t+1})||P_\theta(h_{t+1}))\right]
 \end{align}
}
The sum of two items in the reconstruction indicates that the synthesis of $\mathbf{x}_{t+1}$ involves two steps: (1) draw $h_t$ to generate $s_{t+1}$ and $d_{t+1}$; and (2) estimate $P_\theta(h_{t+1})$ and $P_\phi(h_{t+1})$ based on previously drawn $h_t$, which can be used to derive $h_{t+1}$ and generate $m_{t+1}$.

\paragraph{Multi-generator deliberation for report generation.}
We further generate a medical report $\mathbf{y}_t$ in each visit $t$ based on the medical events $\mathbf{x}_t$, so that the paragraphs of $\mathbf{y}_t$ correspond to different types of events to enhance the informativeness of synthetic EHR data.
The health state $h_t$ is also treated as the conditional signal for report generation.
In order to ensure flexible collaboration between the generation of chief complaint $y_{s_t}$, present illness $y_{d_t}$ and prescription $y_{m_t}$, we employ three generators, each responsible for producing a specific paragraph of $\mathbf{y}_t$.
As these three paragraphs are interdependent, we initially generate $y_{s_t}$, $y_{d_t}$, and $y_{m_t}$ sequentially based on $h_t$ and $\mathbf{x}_t$.
The generative process of $\mathbf{y}_t$ can be formulated as :
{\scriptsize
\begin{align}
  & \log {P}(\mathbf{y}_t | \mathbf{x}_t) \nonumber \\
  = & \ \mathbb{E}_{P_\phi(h_{t\shortminus1})}\left[\log \mathbb{E}_{P_\phi(h_t)} \left( g_s^*(h_t, s_t) g_d^*(h_t, s_t, d_t) g_m^*(h_t, s_t, d_t, m_t)\right) \right] \nonumber \\
  \ge & \ \mathbb{E}_{P_\phi(h_{t\shortminus1})}\big[\mathbb{E}_{P_\phi(h_t)}\log [g_s^*(h_t, s_t) g_d^*(h_t, s_t, d_t) 
  g_m^*(h_t, s_t, d_t, m_t)]
  \nonumber \\
  - & \ \text{KL}(P_\phi(h_t)||P_\theta(h_t))\big], \label{generator}
\end{align}
}
where $g_s^*$, $g_d^*$, and $g_m^*$ are the symptom, diagnosis, and medication report generators, respectively.
Our focus lies in enhancing the collaboration between distinct paragraphs to improve overall generation quality. Consequently, we introduce a multi-generator deliberation framework to facilitate paragraph generation collaboration among multiple generators via message passing and deliberation polish.
As shown in Figure~\ref{fig:model}(c), we assume that report generation consists of a draft phase and a polish phase.
In the draft phase, multiple generators share the same patient's health state $h_t$ as the control signal. 
Additionally, medical events can serve as constraints for generating corresponding paragraphs, thereby improving the alignment between events and reports.
Therefore, we can obtain three draft paragraphs $\widetilde{y}_{s_t}$, $\widetilde{y}_{d_t}$, and $\widetilde{y}_{m_t}$.
To further enhance message passing and collaboration within our multi-generator framework, we draw inspiration from the deliberation networks~\cite{deliberation_Networks1,deliberation_Networks3} and design a refining mechanism in the polish phase to revise the drafts.
Specifically, we encode the draft $\widetilde{y}_{m_t}$ and use the encoded draft $e_{\widetilde{y}_{m_t}}$ as feedback input for the symptom generator and diagnosis generator.
Similarly, for the other two drafts, each is passed as feedback to two other generators different from itself.
By incorporating draft feedback, the framework promotes iterative refinement and collaboration among the generators, leading to the generation of refined paragraphs $\widehat{y}_{s_t}$, $\widehat{y}_{d_t}$, and $\widehat{y}_{m_t}$.

\subsection{Model implementation}
We will detail the implementation of our method in this subsection.
The model overview of \model is illustrated in Figure~\ref{fig:model}, including health state inference, medical event synthesis, and medical report generation. 
\paragraph{Health state inference module.}
We employ three separate encoders to encode the input medical events.
At the $t$-th visit, taking the symptoms $\mathbf{s}_t = (s_{t_1}, \cdots, s_{t_{|\mathbf{s}_t|}})$ as an example, each symptom item is first encoded through an embedding layer. Then, a transformer encoder is used to further process $|\mathbf{s}_t|$ embeddings and integrate them into a symptom representation $e_{s_t}$.
To better model the previous events, we use an attention mechanism $g_{attn}$ to update $e_{s_t}$ into $\widetilde{e}_{s_t}$.
Similarly, the diagnosis and medication encoder can be used to encode corresponding representations $e_{d_t}$ and $e_{m_t}$.
Following the structure of our graphical model, We combine the previously derived $\widetilde{e}_{s_t}$ to get the updated representation $\widetilde{e}_{d_t}$ using two multi-layer perceptions (MLPs) $g_d$ and $g_m$. Similarly, $\widetilde{e}_{m_t}$ can be obtained based on $\widetilde{e}_{s_t}$ and $\widetilde{e}_{d_t}$.
The encoding process can be formulated as:
\begin{align}
    \widetilde{e}_{s_t} & = g_{attn}(e_{s_1}, \cdots, e_{s_t}), \nonumber \\
    \widetilde{e}_{d_t} & = g_d(g_{attn}(e_{d_1}, \cdots, e_{d_t})\oplus\widetilde{e}_{s_t}), \nonumber \\
    \widetilde{e}_{m_t} & = g_{m}(g_{attn}(e_{m_1}, \cdots, e_{m_t})\oplus\widetilde{e}_{s_t}\oplus\widetilde{e}_{d_t}),
\end{align}
where $\oplus$ is the concatenation operation.
The next critical involves modeling the mapping from these encoded representations to the variational parameters for the latent health state $h_t$.
To perform sequential health state inference, we estimate the prior and posterior distributions of $h_t$ using the prior network $P_\theta$ and the inference network $P_\phi$, respectively. 
For the prior network, we first incorporate the input representations $\widetilde{e}_{s_t}$ and $\widetilde{e}_{d_t}$ to infer the distribution $P_\theta(h_t)\sim \mathcal{N}(\mu_{\theta_t},\Sigma_{\theta_t})$.
The prior variables $\mu_{\theta_t}$
and $\log \Sigma_{\theta_t}$ are then estimated via two fully connected layers $g_{\mu_{\theta_t}}$ and $g_{\Sigma_{\theta_t}}$.
For the posterior network, the input representations include an additional  $\widetilde{e}_{m_t}$ to infer $P_\phi(h_t)\sim \mathcal{N}(\mu_{\phi_t},\Sigma_{\phi_t})$, and posterior variables $\mu_{\phi_t}$ and $\Sigma_{\phi_t}$ are similarly estimated via $g_{\mu_{\phi_t}}$ and $g_{\Sigma_{\phi_t}}$.
Thus, the prior and inference networks can be specified as:
{\footnotesize
\begin{align}
    \mu_{\theta_t} = g_{\mu_{\theta_t}}(\widetilde{e}_{s_t}\oplus \widetilde{e}_{d_t}),\quad & 
    \mu_{\phi_t} = g_{\mu_{\phi_t}}(\widetilde{e}_{s_t}\oplus \widetilde{e}_{d_t}\oplus \widetilde{e}_{m_t})
    , \\
    \Sigma_{\theta_t} = g_{\Sigma_{\theta_t}}(\widetilde{e}_{s_t}\oplus \widetilde{e}_{d_t})
    ,\quad & \Sigma_{\phi_t} = g_{\Sigma_{\phi_t}}(\widetilde{e}_{s_t}\oplus \widetilde{e}_{d_t}\oplus \widetilde{e}_{m_t}).
\end{align}
}
With the re-parameterization trick, we can draw a latent variable $z_t$ from $P_\theta(h_t)$ or $P_\phi(h_t)$.
To model the impact of previous health state $h_{t-1}$ on $h_t$, we further refine $z_t$ with an attention mechanism $f_{attn}$ to derive the final health state:
\begin{equation}
    h_t=f_{attn}(z_1, \cdots,z_{t-1}, z_t)
\end{equation}

\paragraph{Medical event synthesis module.}
To make cross-type and longitudinal synthesis for different medical events, we use three separate decoders to generate symptoms, diagnoses, and medications, respectively.
Based on the inferred health state $h_t$, we can generate the current $m_t$ as well as $d_{t+1}$ and $s_{t+1}$ for the next visit.
Specifically, we use three MLPs $f_{\theta_s}$, $f_{\theta_d}$ and $f_{\theta_m}$ with a sigmoid function $\sigma(\cdot)$ to predict the probabilities of each item in these events.
Finally, the predicted probabilities $\widehat{\mathbf{m}}_{t}$, $\widehat{\mathbf{s}}_{t+1}$, $\widehat{\mathbf{d}}_{t+1}$ can be derived as:
{\footnotesize
\begin{align}
    \widehat{\mathbf{m}}_{t} & = \sigma(f_{\theta_m}(h_t \oplus \widetilde{e}_{s_t}\oplus \widetilde{e}_{d_t})), \\
    \widehat{\mathbf{s}}_{t+1}& = \sigma(f_{\theta_s}(h_t)), \\
    \widehat{\mathbf{d}}_{t+1}
    & = \sigma(f_{\theta_d}(h_t \oplus \widetilde{e}_{s_{t+1}})).
\end{align}
}

\paragraph{Medical report generation module.}
Following the recent progress on medical text generation~\cite{biobart}, we adopt BioBART as the backbone of our generators to generate various paragraphs of the report $\mathbf{y}_t$ based on $\mathbf{x}_t$ and $h_t$.
We treat the health state $h_t$ the learned representations of $\mathbf{x}_t$ as continuous prompts and input them into the BART encoder, and then use the BART decoder to generate corresponding report paragraphs.
Following the Fixed-Prompt LM-Tuning strategy~\cite{fixed2,fixed1}, we only unfreeze the BART decoder parameters to optimize the generation performance.
More importantly, we consider enhancing message passing among the paragraphs by two-phase decoding using a multi-generator deliberation framework.
In the draft phase, we sequentially generate three paragraphs $y_{s_t}$, $y_{d_t}$ and $y_{m_t}$ conditioned on $h_t$ and the corresponding event representations. Then, we re-feed these drafts into the BART encoder to obtain its paragraph-level representations $e_{y_{s_t}}$, $e_{y_{d_t}}$ and $e_{y_{m_t}}$. Thus, during the polish phase, we can leverage the feedback information from subsequent generators to refine the generation performance of previous generators.
For instance, $e_{y_{d_t}}$ and $e_{y_{m_t}}$ are used as new continuous prompts to generate a new paragraph $\widetilde{y}_{s_t}$. Similarly, for the other two drafts, each of them is passed as feedback to refine the other two paragraphs.

\begin{table*}
  \centering
  \caption{Medical event synthesis results of unigram, same record \& sequential visit bigram probabilities, and dim-wise fidelity.}
  \resizebox{0.82\textwidth}{!}{
    \begin{tabular}{@{}lcccc|cccc@{}}
    \toprule
    \textbf{Dataset} & \multicolumn{4}{c|}{\textbf{MIMIC-III}} & \multicolumn{4}{c}{\textbf{MIMIC-IV}} \\
    \midrule
    \textbf{Method} & \textbf{Unigram} & \textbf{Bigram} & \textbf{Seq Bigram} & \multicolumn{1}{l|}{\textbf{DimWise}} & \textbf{Unigram} & \textbf{Bigram} & \textbf{Seq Bigram} & \multicolumn{1}{l}{\textbf{DimWise}} \\
    \midrule
    LSTM+MLP & -0.441 & -2.880  & -3.222  & 0.448  & 0.316  & -0.789  & -1.061  & 0.642 \\
    MedGAN & 0.115  & -2.214  & -2.211  & 0.804  & 0.537  & -0.436  & -0.178  & 0.801  \\
    MedBGAN & -0.488  & -5.909  & -6.001  & 0.859  & 0.450  & -0.374  & -0.202  & 0.838  \\
    MedWGAN & 0.742  & 0.386  & 0.425  & 0.813  & 0.771  & 0.587  & 0.581  & 0.857  \\
    EVA   & 0.077  & -5.104  & -2.644  & 0.739  & 0.532  & -1.390  & -0.502  & 0.785  \\
    MTGAN  & 0.804 & 0.608 & 0.597 & 0.869 & 0.790 & 0.569 & 0.591 & 0.875\\
    PromptEHR & 0.795  & 0.616  & 0.603  & 0.921  & 0.775  & 0.600  & 0.588  & 0.855  \\
    \midrule
    \model w/o $h_t$ & 0.605  & 0.169  & -0.015  & 0.872  & 0.529    & 0.115    & 0.001 & 0.860 \\
    \model w/o MV & 0.820  & 0.641  & 0.552  & 0.897  & 0.826    & 0.670    & 0.643    & 0.887 \\
    \textbf{\model} & \textbf{0.846} & \textbf{0.701} & \textbf{0.645} & \textbf{0.972} & \textbf{0.838} & \textbf{0.684} & \textbf{0.648} & \textbf{0.956} \\
    \bottomrule
    \end{tabular}}
  \label{tab:similarity}
\end{table*}

\subsection{Model learning}
We optimize the model parameters by applying a neural variational inference algorithm.
For a specific patient $j$ at visit $t$, we can compute the reconstructed probabilities for each type of events $\widehat{\mathbf{s}}_t^{(j)}$, $\widehat{\mathbf{d}}_t^{(j)}$, and $\widehat{\mathbf{m}}_t^{(j)}$ based on the medical event synthesis module.
We denote $\mathcal{L}_{s_t}^{(j)}$, $\mathcal{L}_{d_t}^{(j)}$, and $\mathcal{L}_{m_t}^{(j)}$ as the binary cross-entropy between generated events and ground-truth events, detailed in Appendix.
Next, we can reconstruct the report $\widehat{\mathbf{y}}_t^{(j)}$ based on the medical report generation module.
We denote $\mathcal{L}_{y_{s_t}}^{(j)}$, $\mathcal{L}_{y_{d_t}}^{(j)}$, and $\mathcal{L}_{y_{m_t}}^{(j)}$ as the language modeling loss between generated and ground-truth paragraphs, also detailed in Appendix.
Then the overall training loss $\mathcal{L}$ can be given by:
{\footnotesize
\begin{align}
    \mathcal{L}_{x_t}^{(j)} & = \lambda_s \mathcal{L}_{s_t}^{(j)} + \lambda_d \mathcal{L}_{d_t}^{(j)} + \lambda_m \mathcal{L}_{m_t}^{(j)}, \\
    \mathcal{L}_{y_t}^{(j)} & = \lambda_{y_s} \mathcal{L}_{y_{s_t}}^{(j)} + \lambda_{y_d} \mathcal{L}_{y_{d_t}}^{(j)} + \lambda_{y_m} \mathcal{L}_{y_{m_t}}^{(j)}, \\
    \mathcal{L}_t^{(j)} & = \mathcal{L}_{x_t}^{(j)} + \mathcal{L}_{y_t}^{(j)} + \lambda_{kl} \text{KL}(P_\phi(h_t) || P_\theta(h_t)),\\
    \mathcal{L} & = \sum_{j=1}^{N}\sum_{t=1}^{T_j}\mathcal{L}_t^{(j)},
\end{align}
}
where $\lambda_*$ are hyper-parameters for the weights of each item.
The reconstruction term attempts to bring synthetic data close to real data and the KL term aims to close the distance between the posterior and prior of the health state.
The training process of \model can be summarized as shown in Algorithm 1 in Appendix.

\section{Experiments}

\subsection{Experimental setup}
\paragraph{Datasets.}
We use real-world EHR data from publicly available datasets MIMIC-III~\cite{MIMIC-3} and MIMIC-IV~\cite{MIMIC-4}. 
Following the data processing in~\cite{drug_recommendation1}, we can directly obtain the diagnosis and medication codes as their labels, and extract symptoms from the clinical notes.
We collect the three most important paragraphs of chief complaint, present illness and prescription from the discharge summary, and desensitize the patient information to create gold medical reports.
The statistics of these two datasets are available in Appendix.
\footnote{Our code and data are available at \url{ https://github.com/p1nksnow/MSIC/}.}

\paragraph{Baselines.}
We compare \model to mainstream synthetic EHR generation methods as baselines.
For medical event synthesis, the baselines include: (1) \textbf{LSTM+MLP} is a straightforward sequential prediction model to yield event probabilities across multiple visits;
(2) \textbf{MedGAN}~\cite{medGAN} is a pioneer in using GAN for synthetic EHR generation; (3) \textbf{MedBGAN} \& \textbf{MedWGAN}~\cite{MedWGANMedBGAN} take advantage of Boundary-seeking GAN~\cite{bgan} and Wasserstein GAN~\cite{wgan} to improve the generation performance; (4) \textbf{EVA}~\cite{EVA} is a VAE-based model featuring a BiLSTM encoder and CNN decoder; (5) \textbf{MTGAN}~\cite{mtgan} is a time-series generation model via conditional GAN; and (6) \textbf{PromptEHR}~\cite{promptehr} employs prompt learning for conditional EHR generation.

Medical report generation is our first proposal and we select some representative pre-trained language models as baselines: (1) \textbf{Transformer}~\cite{Transformer}, the original vanilla architecture for sequence-to-sequence generation; (2) \textbf{GPT-2}~\cite{GPT-2} is a decoder-only auto-regressive language model pre-trained on open domain corpora; (3) \textbf{BioGPT}~\cite{biogpt} is recently pre-trained on large-scale biomedical literature; (4) \textbf{BART}~\cite{bart} is an encoder-decoder model pre-trained in the general domain; and (5) \textbf{BioBART}~\cite{biobart} adapts BART to the biomedical domain.

\paragraph{Evaluation metrics.}
To assess the quality of medical event synthesis, we leverage the following metrics to measure the reliability of synthetic EHR data.
\textbf{Statistical similarity}\footnote{
This metric is adopted by the official function ``r2\_score'' in ``sklearn.metrics''.
Lower (even negative) values indicate greater negative correlations.
}: We count the frequency for each event in real and synthetic data and compute the coefficient of determination for the two frequency lists, yielding the \textbf{Unigram} score.
Besides, we consider the co-occurrence frequency of each event pair in the same visit to compute the \textbf{Bigram} score, or across two consecutive visits for the \textbf{Seq Bigram} score.
\textbf{Fidelity}: We compute the correlation coefficient of the dimension-wise distributions of each event in real and synthetic records as the \textbf{Dimwise} score.
For medical report generation, we use text generation-based metrics including \textbf{BLEU}~\cite{BLEU}  and \textbf{ROUGE}~\cite{ROUGE} to evaluate n-gram overlap between generated and gold reports.

\subsection{Quality of synthetic data}
\paragraph{Medical event synthesis.}
As shown in Table~\ref{tab:similarity},
MedGAN and EVA perform poorly on these metrics, indicating that original generative models cannot capture sufficient information from real records.
MedBGAN and MedWGAN slightly improve the performance with new GAN-style training. 
MTGAN and PromptEHR can synthesize better records based on conditional generation patterns.
\model can outperform all baselines on both statistical similarity and fidelity, indicating that our graphical model can help improve the quality of synthetic EHR data.
Limited by space, we only show the results of combining all types of events, and more details are included in Appendix.

\begin{table}[t]
  \centering
  \caption{Report generation comparison on MIMIC-III.}
  \resizebox{0.85\linewidth}{!}{
    \begin{tabular}{@{}lcccc@{}}
    \toprule
    \textbf{Method} & \textbf{BLEU-2} & \textbf{ROUGE-L} & \textbf{BLEU-2} & \textbf{ROUGE-L} \\
    \midrule
    Transformer & 21.63 & 9.21  & 16.66 & 13.99 \\
    GPT-2 & 22.17  & 13.95  & 19.96  & 9.32  \\
    BioGPT & 27.26  & 23.03  & 26.57  & 22.42  \\
    BART  & 27.00  & 20.78  & 19.13  & 20.49  \\
    BioBART & 29.80  & 24.39  & 28.57  & 27.22  \\
    \midrule
    \model w/o $h_t$ & 31.17  & 26.56  & 28.72  & 26.97  \\
    \model w/o MV & 32.94  & 27.55  & 28.96  & 26.92  \\
    \model w/o MA & 29.51  & 25.13  & 26.93  & 25.28  \\
    \textbf{\model} & \textbf{33.23} & \textbf{27.79} & \textbf{29.64} & \textbf{27.41} \\
    \bottomrule
    \end{tabular}}
  \label{tab:report}
\end{table}

\paragraph{Medical report generation.}
The comparison results of medical report generation are shown in Table~\ref{tab:report}. 
The performance of Transformer is not as good as the pre-trained models.
Since BioGPT and BioBART are pre-trained on the biomedical-domain corpora, they can generate better reports than GPT-2 and BART pre-trained on the open-domain datasets.
\model outperforms the best baseline BioBART through enhanced message passing among multiple generators and two-phase decoding in our deliberation framework. 
More results are available in Appendix.

\paragraph{Ablation study.}
To verify the effectiveness of the individual components in \model, we compare \model with several model variants for ablation analysis: (1) \textbf{\model w/o $h_t$} removes the health state and all paths related to it in Figure~\ref{fig:g1};
(2) \textbf{\model w/o MV} removes the gray arrows in Figure~\ref{fig:g1} without sequential impact across visits;
(3) \textbf{\model w/o MA} removes the message passing between multiple generators and generates different paragraphs independently.
The ablation results in Tables~\ref{tab:similarity} and~\ref{tab:report} show that without modeling the patient's health state, the performance of \model w/o $h_t$ drops a lot, indicating the health state is indeed a key conditional signal for EHR generation. \model w/o MV generating records for a single visit is also inferior to \model, suggesting that multi-visit inference is necessary for coherent records.
\model w/o MA performs worst in report generation, showing the effectiveness of our multi-generator network.

\begin{figure}
\centering
\subfigure[Jaccard.]{
\label{fig:ujac}
\includegraphics[width=0.22\textwidth]{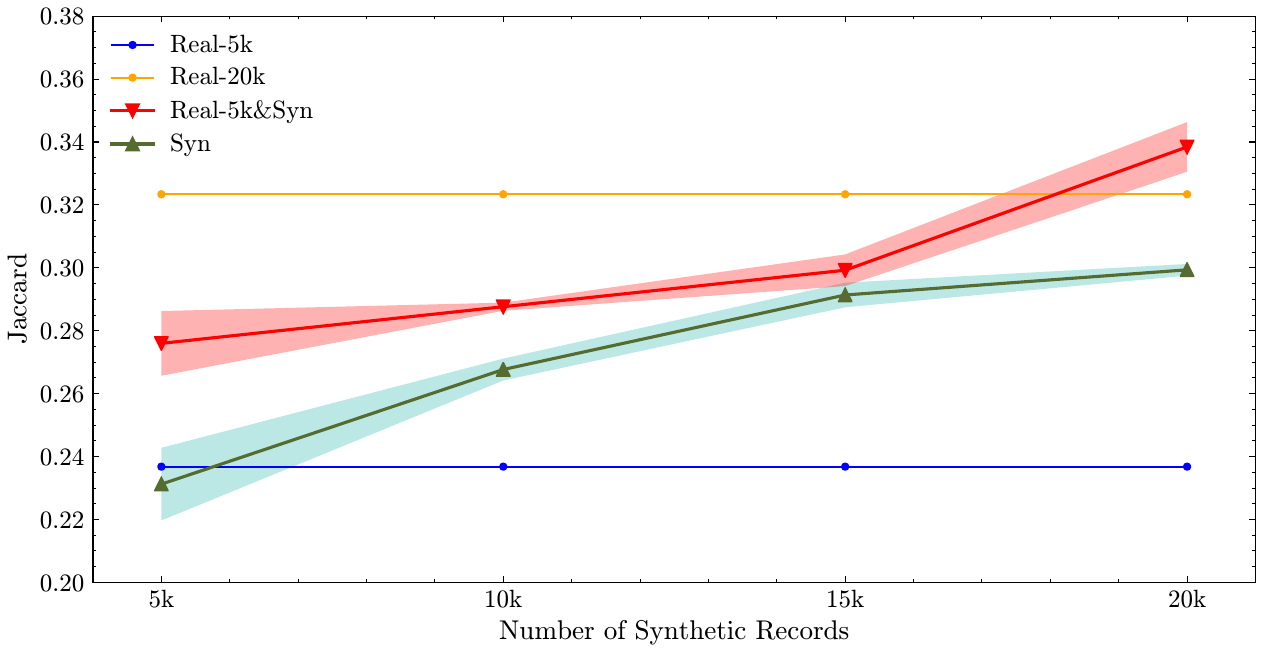}}
\subfigure[F1 score.]{
\label{fig:uf1}
\includegraphics[width=0.22\textwidth]{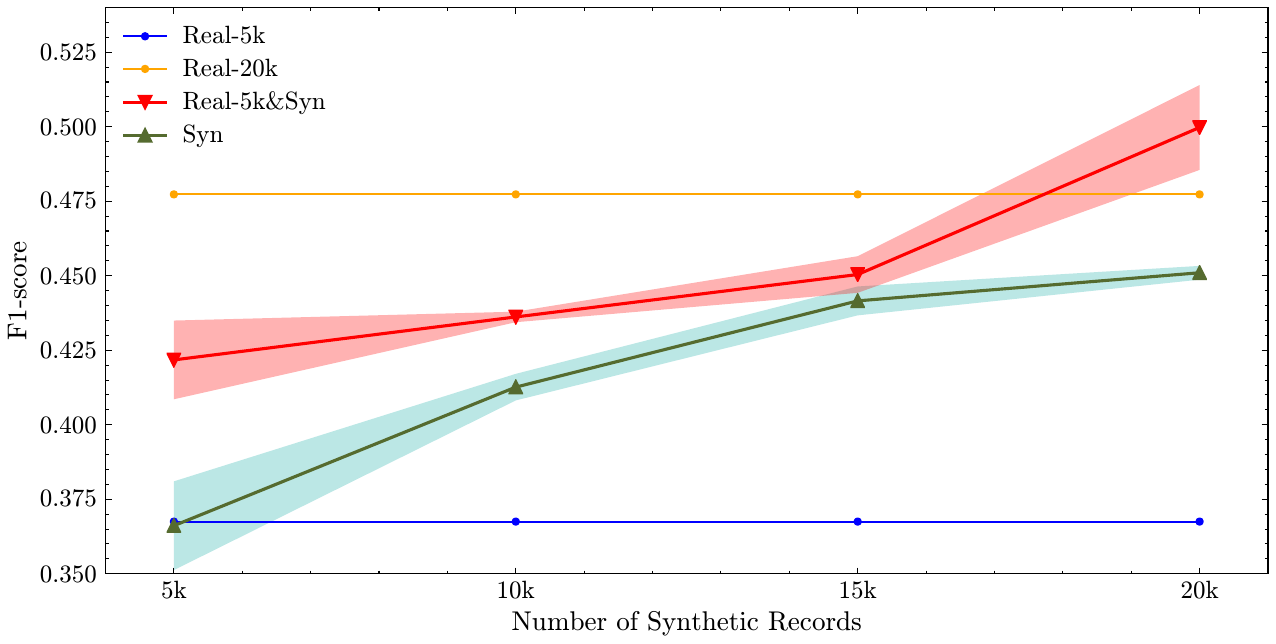}}
\caption{Utility of real, synthetic, and hybrid training data.}
\label{fig:utility}
\end{figure}

\subsection{Utility of synthetic data}
We also measure the utility of synthetic data by comparing the performance of synthetic, real, and hybrid EHR training data.
Specifically, we first use various sizes of fully synthetic records for training and compare them with those trained on 5k and 20k fully real records.
Next, we create hybrid training data by combining different proportions of synthetic records with 5k real records.
Figure~\ref{fig:utility} indicates that the performance using fully synthetic data lies between real 5k and real 20k. As the size of the synthetic data increases, it performs better than the low-resourced real data, but not as well as real data of the same size. Furthermore, large-size hybrid data can break the upper limit of real data, indicating that 
by augmenting real data with synthetic data, it can further improve the utility of patient records.

\begin{figure}
\centering
\subfigure[Membership attack.]{
\label{fig:member}
\includegraphics[width=0.22\textwidth]{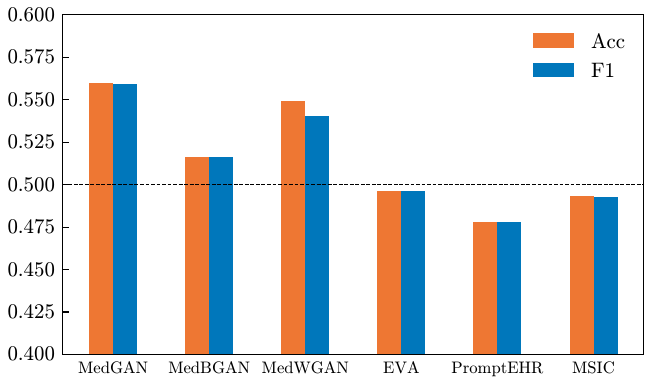}}
\subfigure[Attribute attack.]{
\label{fig:attr}
\includegraphics[width=0.22\textwidth]{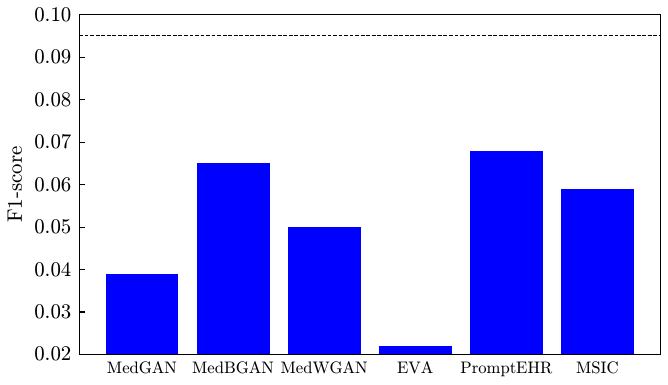}}
\caption{Privacy risk evaluation results on MIMIC-III.}
\label{fig:privacy}
\end{figure}

\vspace{-1mm}
\subsection{Privacy risk evaluation}
We also evaluate the privacy risk of \model compared to the baselines by conducting the following two privacy attacks.

\vspace{-1mm}
\paragraph{Membership inference attack.} 
We follow the evaluation of~\cite{member} for this analysis, which aims to ascertain whether real records similar to synthetic records belong to the training data.
We randomly sample 500 records from the training and test sets respectively, and use the training set to train each model to synthesize records.
For each real record, Jaccard is used as a distance measure to identify the closest synthetic record and classify it as either training or test data.
The ideal value for metrics is random guessing (i.e. 0.5), with higher scores indicating a greater degree of privacy leakage.
The results in Figure~\ref{fig:privacy}(a) show that the metrics of our method are closest to 0.5, indicating enhanced protection against inferring training data from synthetic data.
\vspace{-1mm}
\paragraph{Attribute inference attack.}
We select the most frequently encountered events as common attributes and the rest as sensitive attributes. We identify the closest synthetic patient to each one in the training set, based on the overlap in common attributes. Predicted sensitive attributes are identical to those of the closet synthetic patient, and are evaluated to actual sensitive attributes using the F1 Score. ``Real data attack'' serves as an acceptable attribute inference, using real test records instead of synthetic patients. 
The results in Figure~\ref{fig:privacy}(b)
show that our \model achieves an F1 score of 0.059, lower than Real Data Attack (horizontal line at 0.095). This indicates that our model can withstand this attack without revealing sensitive attributes of real patient data. MedGAN and EVA exhibit much lower scores due to the lack of pattern acquisition from real data. In contrast, \model can capture more sufficient information with low privacy risks.

\vspace{-2mm}
\section{Conclusion}
In this work, we propose \model, a \fullmodel. 
First, we construct a graphical model to capture logical associations between different types of medical events in synthetic EHR generation.
We introduce a latent health state variable for each patient as a key conditional signal for generating different events. 
To improve the coherence of generated records within multiple visits, we extend the health state inference method to the multi-visit scenario to effectively utilize previous records.
Furthermore, we propose a new medical report generation task to provide broader applications for synthetic data.
Our proposed multi-generator deliberation framework enhances collaboration between different generators, and the two-phase decoding strategy facilitates message passing between generators and improves the overall performance of EHR synthesis.

\vspace{-2mm}
\section*{Acknowledgements}
This work was supported by the National Natural Science Foundation of China (NSFC Grant No. 62122089),
Beijing Outstanding Young Scientist Program NO. BJJWZYJH012019100020098, and Intelligent Social Governance Platform, Major Innovation \& Planning Interdisciplinary Platform for the “Double-First Class” Initiative, Renmin University of China, the Fundamental Research Funds for the Central Universities, and the Research Funds of Renmin University of China.

\vspace{-2mm}
\bibliography{aaai24}
\appendix
\onecolumn
\section{Details of Model Training}

\subsection{Training loss}
In the training process, our objective is to minimize the ELBO-based loss, which consists of a reconstruction term to ensure the synthetic records closely resemble the real records, and a KL divergence term to minimize the gap between the prior and posterior distributions.
To reconstruct medical events for each visit $t$ of each patient $j$, we leverage our medical event synthesis module to obtain generated events, and then minimize the binary cross-entropy between the generated and ground-truth events, which can be formulated as:
\begin{align}
    \mathcal{L}_{s_t}^{(j)} & = - [\mathbf{s}_t^{(j)} \log(\widehat{\mathbf{s}}_t^{(j)}) + (1-\mathbf{s}_t^{(j)}) \log(1-\widehat{\mathbf{s}}_t^{(j)})], \nonumber \\
    \mathcal{L}_{d_t}^{(j)} & = - [\mathbf{d}_t^{(j)} \log(\widehat{\mathbf{d}}_t^{(j)}) + (1-\mathbf{d}_t^{(j)}) \log(1-\widehat{\mathbf{d}}_t^{(j)})], \nonumber \\
    \mathcal{L}_{m_t}^{(j)} & = - [\mathbf{m}_t^{(j)} \log(\widehat{\mathbf{m}}_t^{(j)}) + (1-\mathbf{m}_t^{(j)}) \log(1-\widehat{\mathbf{m}}_t^{(j)})]. \nonumber
\end{align}
For the reconstruction of medical reports, we minimize the following language modeling loss:
\begin{align}
    \mathcal{L}_{y_{s_t}}^{(j)} & = -\sum_{k}\log P(y_{s_t, k} | y_{s_t, <k}, \mathbf{x}_{t-1}), \nonumber \\ 
    \mathcal{L}_{y_{d_t}}^{(j)} & = -\sum_{k}\log P(y_{d_t, k} | y_{d_t, <k}, \mathbf{x}_{t-1}, s_t), \nonumber \\ 
    \mathcal{L}_{y_{m_t}}^{(j)} & = -\sum_{k}\log P(y_{m_t, k} | y_{m_t, <k}, \mathbf{x}_{t-1}, s_t, d_t), \nonumber
\end{align}
where $y_{*_t, k}$ represents the $k$-th token of the corresponding paragraph $y_{s_t}$, and $y_{d_t}$ and $y_{m_t}$ follow the same pattern. The predicted probabilities can be estimated using our report generation module. 

\subsection{Training algorithm}
To detail the model training process, our \model is optimized based on the Algorithm~\ref{alg:algorithm}. Note that all equations quoted below are from the main paper. While for the model inference process, the only difference is that the latent variable $z_t^{(i)}$ needs to be drawn from the prior distribution $\mathcal{N}(\mu_{\theta_t}^{(j)}, \Sigma_{\theta_t}^{(j)})$ (line 8). The inference process concludes with the report generation stage (line 13), without updating any network parameters.

\begin{algorithm}
\caption{Training Process for \model}
\label{alg:algorithm}
\textbf{Input}: EHR Training data: $\{\mathbf{x}^{(j)},\mathbf{y}^{(j)}\}_{j=1}^{N}$, number of epoches: $I$, weights of loss functions: $\lambda_{*}$.  \\
\textbf{Parameter}: Learnable network parameters. \\
\textbf{Output}: Synthesized EHR events and reports $\{\widehat{\mathbf{x}}^{(j)},\widehat{\mathbf{y}}^{(j)}\}_{j=1}^{N}$.
\begin{algorithmic}[1] 
\FOR{$i = 1$ to $I$}
\FOR{$j = 1$ to $N$}
\STATE $\mathcal{L} \leftarrow 0$.
\FOR{$t = 1$ to $T_j$}
\STATE // Latent health state inference
\STATE Compute the input representations $\widetilde{e}_{s_t}^{(j)}$, $\widetilde{e}_{d_t}^{(j)}$, and $\widetilde{e}_{m_t}^{(j)}$ using Eq.(6).
\STATE Compute prior/posterior means and log-variances $\mu_{\theta_t}^{(j)}$, $\mu_{\phi_t}^{(j)}$, $\Sigma_{\theta_t}^{(j)}$, and $\Sigma_{\phi_t}^{(j)}$ using Eq.(7)-(8).
\STATE Draw $z_t^{(i)} \sim \mathcal{N}(\mu_{\phi_t}^{(j)}, \Sigma_{\phi_t}^{(j)})$, and derive the health state using Eq.(9).
\STATE // Medical event synthesis
\STATE Predict the synthesized probabilities $\widehat{\mathbf{s}}_{t}$, $\widehat{\mathbf{d}}_{t}$, and $\widehat{\mathbf{m}}_{t}$ for medical events using Eq.(10)-(12).
\STATE // Medical report generation
\STATE Generate three drafted paragraphs sequentially: \\$\widetilde{y}_{s_t} = \text{BART}(h_t, \widetilde{e}_{s_t})$,  $\widetilde{y}_{d_t} = \text{BART}(h_t, \widetilde{e}_{s_t}, \widetilde{e}_{d_t})$, 
$\widetilde{y}_{m_t} = \text{BART}(h_t, \widetilde{e}_{s_t}, \widetilde{e}_{d_t}, \widetilde{e}_{m_t})$.
\STATE Polish three paragraphs with message passing:\\
$\widehat{y}_{s_t} = \text{BART}(h_t,e_{\widetilde{y}_{d_t}}, e_{\widetilde{y}_{m_t}})$, 
$\widehat{y}_{d_t} = \text{BART}(h_t, e_{\widetilde{y}_{s_t}}, e_{\widetilde{y}_{m_t}})$, 
$\widehat{y}_{m_t} = \text{BART}(h_t, e_{\widetilde{y}_{s_t}}, e_{\widetilde{y}_{d_t}})$.
\STATE Compute the loss term $\mathcal{L}_t^{(j)}$ using Eq.(13)-(15).
\STATE $\mathcal{L} \leftarrow \mathcal{L} + \mathcal{L}_t^{(j)}$.
\ENDFOR
\STATE Update the learnable parameters to minimize $\mathcal{L}$.
\ENDFOR
\ENDFOR
\STATE \textbf{return} $\{\widehat{\mathbf{x}}^{(j)},\widehat{\mathbf{y}}^{(j)}\}_{j=1}^{N}$.
\end{algorithmic}
\end{algorithm}

\section{Experiment Details}
\subsection{Dataset details}
Similar to the preprocessing of previous work on MIMIC datasets~\cite{promptehr,drug_recommendation1}, we identify symptoms, diagnoses, and medications as the most representative and applicable medical events for cross-type and longitudinal event synthesis.
In addition, 
we further collect the medical reports from the discharge summary of patients for the new medical report generation task.
The overall statistics of our used EHR datasets are shown in Table \ref{tab:stat}.

\begin{table*}[h]
  \centering
  \caption{The statistics of EHR datasets: MIMIC-III and MIMIC-IV.}
    \begin{tabular}{lcc}
    \toprule
    \textbf{Items} & \textbf{MIMIC-III} & \textbf{MIMIC-IV} \\
    \midrule
    \# of patients & 5794  & 36615 \\
    \# of clinical visits & 12220 & 126134 \\
    sypt./ diag./ med. vocab size & 407/1831/112 & 555/2000/154 \\
    avg. \# of visits & 2.11  & 3.45  \\
    avg. \# of sypt./ diag./ med. events per visit & 7.91/14.20/20.10 & 11.01/12.76/13.31 \\
    avg. tokens of sypt./ diag./ med. reports per visit & 9.15/337.11/296.12 & 6.89/169.22/260.78 \\
    \bottomrule
    \end{tabular}%
  \label{tab:stat}
\end{table*}

\subsection{Implementation details}
For both datasets of MIMIC-III and MIMIC-IV, we randomly split the dataset into train, valid, and test set by different patients with a ratio of 8:1:1.
We utilize the Adam optimizer and a linear warm-up and decreasing learning rate scheduler, initializing the base learning rate at 1e-5. 
The dimensions of the embedding, transformer encoder layer, and hidden states are all set to 256.
For the weights of loss functions, $\lambda_s$, $\lambda_d$, $\lambda_m$ and $\lambda_{y_d}$ are all set to 1.
$\lambda_{y_s}$, $\lambda_{y_m}$, and $\lambda_{kl}$ are set to 5, 2, 1e-4, respectively. 
The model training process lasts for 20 epochs.
We perform only medical event synthesis for the first five epochs and jointly train all modules for event synthesis and report generation in the remaining epochs.

\section{Supplementary Experimental Results}

\subsection{Report generation}
The experimental results of medical report generation on MIMIC-III and MIMIC-IV are shown in Tables~\ref{tab:report3} and~\ref{tab:report4}, which contain more evaluation metrics.
Compared to all baselines and ablation variants, our \model achieves the best performance on almost all metrics.

\begin{table*}[h]
  \centering
  \caption{The experimental results of medical report generation on MIMIC-III.}
  \begin{tabular}{@{}lccccccc@{}}
    \toprule
    \textbf{Method} & \textbf{BLEU-1} & \textbf{BLEU-2} & \textbf{BLEU-3} & \textbf{BLEU-4} & \textbf{ROUGE-1} & \textbf{ROUGE-2} & \textbf{ROUGE-L} \\
    \midrule
    Transformer & 36.95 & 21.63 & 12.62 & 7.35  & 9.29  & 2.21  & 9.21 \\
    GPT-2 & 37.26  & 22.17  & 13.28  & 8.05  & 14.17  & 3.45  & 13.95  \\
    BioGPT & 43.53  & 27.26  & 17.41  & 11.44  & 24.10  & 8.61  & 23.03  \\
    BART  & 40.63  & 27.00  & 18.57  & 13.23  & 21.26  & 9.75  & 20.78  \\
    BioBART & 44.13  & 29.80  & 20.93  & 15.31  & 24.90  & \textbf{11.54} & 24.39  \\
    \midrule
    \model w/o $h_t$ & 47.15  & 31.17  & 21.37  & 15.28  & 27.59  & 9.64  & 26.56  \\
    \model w/o MV & 49.60  & 32.94  & 22.65  & 16.22  & 28.55  & 9.91  & 27.55  \\
    \model w/o MA & 44.60  & 29.51  & 20.22  & 14.43  & 25.90  & 8.83  & 25.13  \\
    \textbf{\model} & \textbf{50.05} & \textbf{33.23} & \textbf{22.86} & \textbf{16.38} & \textbf{28.81} & 10.00  & \textbf{27.79} \\
    \bottomrule
    \end{tabular}
  \label{tab:report3}
\end{table*}

\begin{table*}[h]
  \centering
  \caption{The experimental results of medical report generation on MIMIC-IV.}
    \begin{tabular}{@{}lccccccc@{}}
    \toprule
         \textbf{Method} & \textbf{BLEU-1} & \textbf{BLEU-2} & \textbf{BLEU-3} & \textbf{BLEU-4} & \textbf{ROUGE-1} & \textbf{ROUGE-2} & \textbf{ROUGE-L} \\
    \midrule
    Transformer & 27.7  & 16.66 & 10.03 & 0.06  & 14.11 & 4.61  & 13.99 \\
    GPT-2 & 34.12  & 19.96  & 11.72  & 6.89  & 9.76  & 2.84  & 9.32  \\
    BioGPT & 42.21  & 26.57  & 16.84  & 10.81  & 22.70  & 8.83  & 22.42  \\
    BART  & 29.79  & 19.13  & 12.28  & 7.99  & 20.84  & 5.96  & 20.49  \\
    BioBART & 44.79  & 28.57  & 18.28  & 11.86  & 27.95  & \textbf{9.44} & 27.22  \\
    \midrule
    \model w/o $h_t$ & 44.84  & 28.72  & 18.62  & 12.28  & 27.93  & 8.51  & 26.97  \\
    \model w/o MV & 45.17  & 28.96  & 18.80  & 12.43  & 28.06  & 8.47  & 26.92  \\
    \model w/o MA & 41.29  & 26.93  & 17.75  & 11.78  & 26.16  & 7.97  & 25.28  \\
    \textbf{\model} & \textbf{45.62} & \textbf{29.64} & \textbf{19.44} & \textbf{13.03} & \textbf{28.54} & 9.19  & \textbf{27.41} \\
    \bottomrule
    \end{tabular}
  \label{tab:report4}
\end{table*}

\subsection{Case study}
In addition to the quantitative results, we also conduct a case study to evaluate the quality of generated synthetic records using different models.
We choose BioBART~\cite{biobart} and \model w/o MA as strong baselines for comparison.
The results show that the report generated by our \model is the closest to the ground-truth report.
We can see erroneous events and duplicate text in the reports generated by BioBART, which affected the synthetic quality.
Without the multi-generator deliberation framework to collaborate multiple generators, the coherence and informativeness of the reports generated by \model w/o MA are also significantly reduced.

However, the limitation of \model is that there are still some false positive events in our synthetic report.
Generating events not appearing in the real data is still a challenge for synthetic EHR generation tasks.
We will refine this problem and aim to further reduce false positives in our future work.
Regardless, \model already has the fewest false positives compared to existing methods.

\begin{table*}[h]
  \centering
  \caption{An example synthetic report for a visit. Matched key events are marked in bold.}
  \resizebox{\textwidth}{!}{
    \begin{tabular}{@{}lll@{}}
    \toprule
    \textbf{Method} & \textbf{Report paragraph} & \textbf{Report content} \\
    \midrule
    \multirow{6}[6]{*}{Ground-truth} & Chief complaint & \textbf{Abdominal pain.} \\
\cmidrule{2-3}          & \multirow{2}[2]{*}{Present illness} & The patient is being admitted with \textbf{abdominal pain}, with labs showing transaminitis and \textbf{pancreatitis}.  \\
          &       & This pain is located in the \textbf{epigastrium}. Patient was prompted to the emergency department by physician. \\
\cmidrule{2-3}          & \multirow{3}[2]{*}{Prescription} & \multicolumn{1}{p{45em}}{1.  \textbf{Acetaminophen 1000 mg PO Q8H:PRN Pain} - Mild/Fever, please limit to 4000mg in 24 hour period.} \\
          &       & \multicolumn{1}{p{45em}}{2.  \textbf{OxyCODONE (Immediate Release) 5 mg PO Q4H:PRN Pain} - Moderate, medication may cause drowsiness.} \\
          &       & 3.  Polyethylene Glycol 17 g PO DAILY:PRN Constipation - discontinue when bowel pattern normalizes. \\
    \midrule
    \multirow{6}[6]{*}{\model} & Chief complaint & \textbf{Abdominal pain.} \\
\cmidrule{2-3}          & \multirow{2}[2]{*}{Present illness} & The patient suffer from \textbf{pancreatitis}, hypertension, recent diagnosed with painful post-stent placement. \\
          &       & Patient was found to have worsening SOB with \textbf{epigastric pain}. \\
\cmidrule{2-3}          & \multirow{3}[2]{*}{Prescription} & \multicolumn{1}
{p{45em}}
{1.  \textbf{Acetaminophen 1000 mg PO Q8H:PRN Pain} - Mild.} \\
          &       & \multicolumn{1}{p{45em}}{2.  Docusate Sodium 100 mg PO BID.} \\
          &       & 3.  \textbf{OxyCODONE (Immediate Release) 5 mg PO Q4H:PRN Pain} - Moderate. \\
    \midrule
    \multirow{6}[6]{*}{\model w/o MA} & Chief complaint & \textbf{Abdominal pain.} \\
\cmidrule{2-3}          & \multirow{2}[2]{*}{Present illness} & The patient has history of HCV cirrhosis and recent complicated by portal hypertension who presents \\
          &       & \textbf{abdominal pain}. This has been worse last several months, but now has been present with a pain/nausea. \\
\cmidrule{2-3}          & \multirow{3}[2]{*}{Prescription} & \multicolumn{1}{p{45em}}{1.  \textbf{Acetaminophen 1000 mg PO Q8H}.} \\
          &       & \multicolumn{1}{p{45em}}{2.  Docusate Sodium 100 mg PO BID.} \\
          &       & 3. Aspirin 81 mg PO DAILY. \\
    \midrule
    \multirow{6}[6]{*}{BioBART} & Chief complaint & Left sided weakness. \\
\cmidrule{2-3}          & \multirow{2}[2]{*}{Present illness} & The patient states he has not able to tell the area of his bowel pain.  Patient denies any trauma. \\
          &       & Patient has significant \textbf{abdominal pain} that the swelling. \\
\cmidrule{2-3}          & \multirow{3}[2]{*}{Prescription} & \multicolumn{1}{p{45em}}{1.  \textbf{Acetaminophen 1000 mg PO Q8H}.} \\
          &       & \multicolumn{1}{p{45em}}{2.  Docusate Sodium 100 mg PO BID.} \\
          &       & 3.  Docusate Sodium 100 mg PO BID. \\
    \bottomrule
    \end{tabular}}
  \label{tab:case}
\end{table*}

\begin{figure*}
\centering
\subfigure[Membership inference attack.]{
\label{fig:member4}
\includegraphics[width=0.44\textwidth]{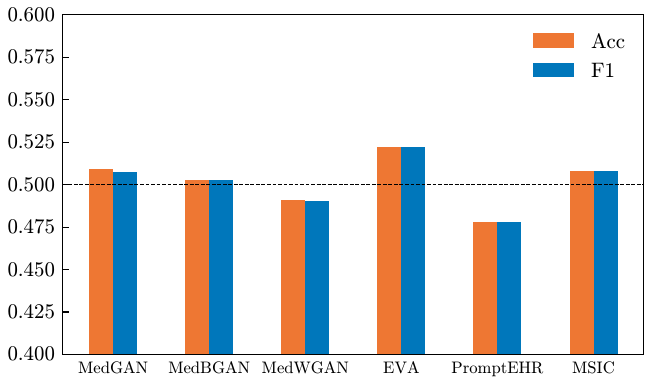}}
\subfigure[Attribute inference attack.]{
\label{fig:attr4}
\includegraphics[width=0.44\textwidth]{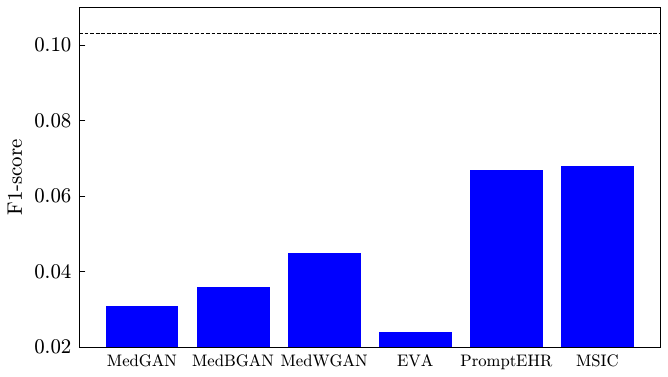}}
\caption{The results of privacy risk evaluation on MIMIC-IV.}
\label{fig:privacy2}
\end{figure*}

\subsection{Privacy risk evaluation}
In addition to the privacy risk evaluation results on MIMIC-III in the main paper, here we also include the membership inference attack and attribute inference attack on MIMIC-IV. The results in Figure~\ref{fig:privacy2} show that for the membership inference attack, the accuracy and F1 score close to 0.5 indicate that our \model can better prevent inferring whether each record from synthetic data exists in the training set.

For the attribute inference attack, the F1 score of \model is 0.068, lower than Real Data Attack (horizontal line at 0.103), indicating that our model can withstand this attack without revealing sensitive attributes of real patient data. Lower F1 scores indicate that fewer patterns are successfully extracted from real data. \model can capture more sufficient information with low privacy risks.

\subsection{Statistical similarity visualization}
\begin{figure*}[h]
    \centering
    \includegraphics[width=\textwidth]{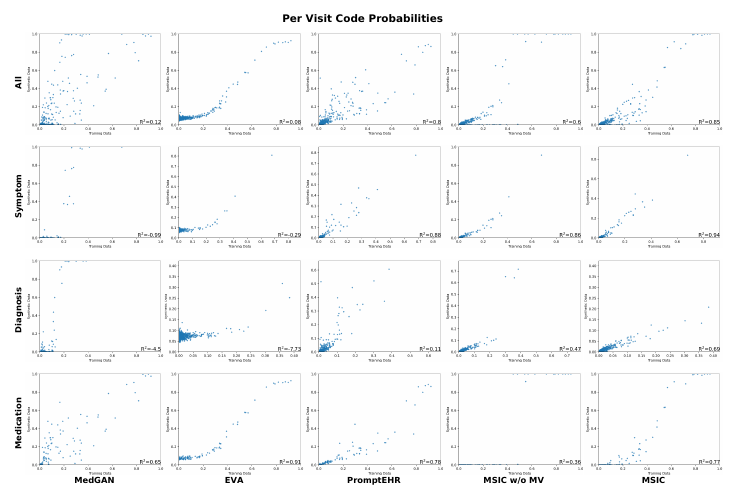}
    \caption{The Unigram probabilities of synthetic events on MIMIC-III.}
    \label{fig:3unigram}
\end{figure*}

\begin{figure*}
    \centering
    \includegraphics[width=\textwidth]{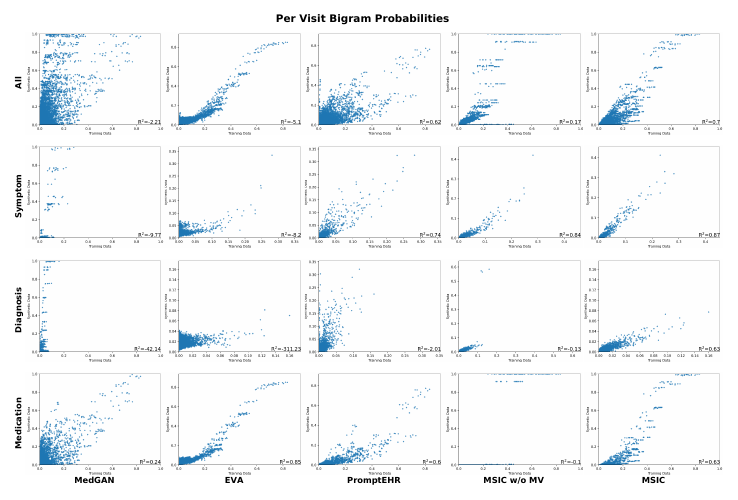}
    \caption{The Bigram probabilities of synthetic events on MIMIC-III.}
    \label{fig:3bigram}
\end{figure*}

\begin{figure*}
    \centering
    \includegraphics[width=\textwidth]{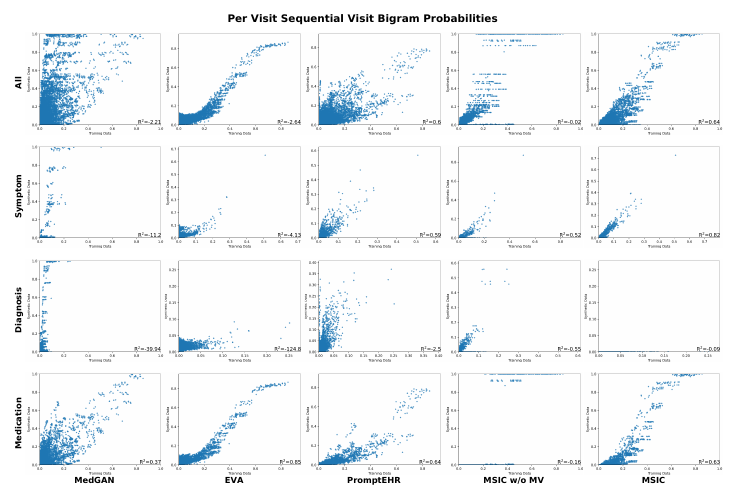}
    \caption{The Sequential Visit Bigram probabilities of synthetic events on MIMIC-III.}
    \label{fig:3seq}
\end{figure*}

\begin{figure*}
    \centering
    \includegraphics[width=\textwidth]{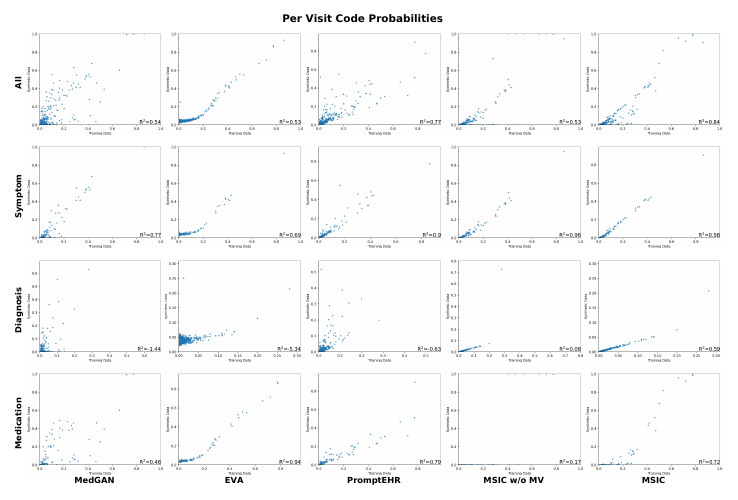}
    \caption{The Unigram probabilities of synthetic events on MIMIC-IV.}
    \label{fig:4unigram}
\end{figure*}

\begin{figure*}
    \centering
    \includegraphics[width=\textwidth]{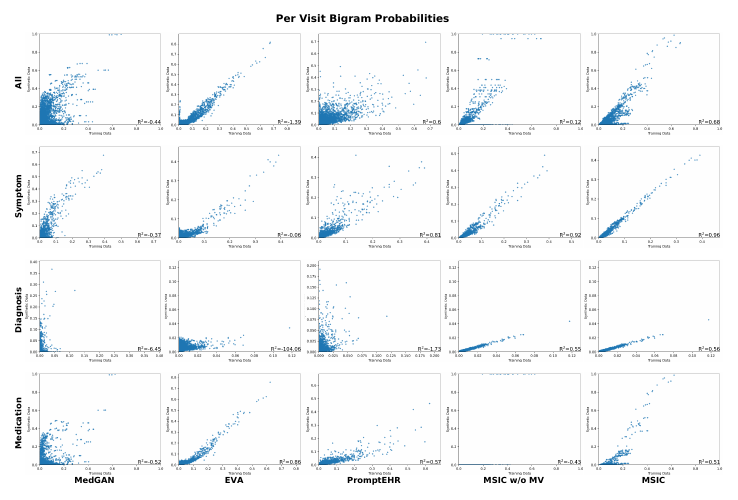}
    \caption{The Bigram probabilities of synthetic events on MIMIC-IV.}
    \label{fig:4bigram}
\end{figure*}

\begin{figure*}
    \centering
    \includegraphics[width=\textwidth]{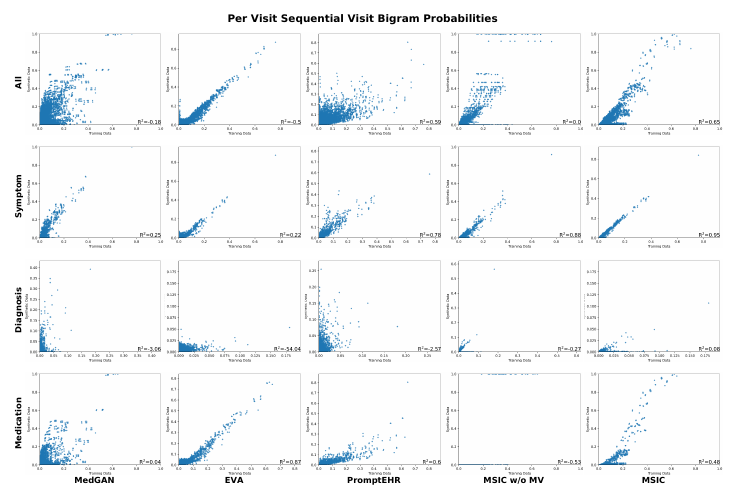}
    \caption{The Sequential Visit Bigram probabilities of synthetic events on MIMIC-IV.}
    \label{fig:4seq}
\end{figure*}

We evaluate the statistical similarity between the real and synthetic EHRs.
We compare our \model with some baselines: PromptEHR~\cite{promptehr}, EVA~\cite{EVA}, MedGAN~\cite{medGAN}) and \model w/o MV.
We first use each method to generate synthetic sets of the same size as the training set for fair comparison.
Then, we calculate the \textbf{Unigram} score using the frequency of each unique event (a.k.a code) in both training data and synthetic data, respectively.
Similarly, we can also compute the Same Record \textbf{Bigram} score and Sequential Visit Bigram (\textbf{Seq Bigram}) score by computing the co-occurrence frequency for each event pair within the same visit or across sequential visits.
Finally, we collect two frequency groups and calculate their correlation coefficients to evaluate the statistical similarity between synthetic data and training data.
For each metric of each method, We visualize the event frequency for each type (Symptom, Diagnosis, Medication) and combined types (All), and the resulting scatter plots are illustrated in Figures~\ref{fig:3unigram}-~\ref{fig:4seq}.
The results show that our \model substantially outperforms other baselines on the statistical similarity of various types of events. This indicates that the distribution of EHR data synthesized by \model is closest to the distribution of real EHR data, and thus it has higher reliability and is more in line with medical knowledge.

\end{document}